\documentclass[10pt, a4paper]{article}

\usepackage[]{lrec-coling2024} 

\usepackage{makecell}
\usepackage{subfig}
\usepackage{amsmath}
\usepackage{amsfonts}
\usepackage{multirow}
\usepackage{dblfloatfix}
\usepackage{scrextend}
\usepackage{url}
\usepackage{adjustbox}
\usepackage{makecell}
\usepackage[most]{tcolorbox}
\usepackage{booktabs}
\usepackage{cuted}

\usepackage[most]{tcolorbox}

\title{Deconstructing In-Context Learning: Understanding Prompts via Corruption\\ 
}

\name{Namrata Shivagunde, Vladislav Lialin, Sherin Muckatira, Anna Rumshisky} 
\address{University of Massachusetts Lowell \\
{\{nshivagu, vlialin, smuckati, arum\}}@cs.uml.edu\\}

\abstract{
The ability of large language models (LLMs) to ``learn in contex'' based on the provided prompt has led to an explosive growth in their use, culminating in the proliferation of AI assistants such as ChatGPT, Claude, and Bard. These AI assistants are known to be robust to minor prompt modifications, mostly due to alignment techniques that use human feedback. In contrast, the underlying pre-trained LLMs they use as a backbone are known to be brittle in this respect. 
Building high-quality backbone models remains a core challenge, and a common approach to assessing their quality is to conduct few-shot evaluation. Such evaluation is notorious for being highly sensitive to minor prompt modifications, as well as the choice of specific in-context examples. 
Prior work has examined how modifying different elements of the prompt can affect model performance. However, these earlier studies tended to concentrate on a limited number of specific prompt attributes and often produced contradictory results.  
Additionally, previous research either focused on models with fewer than 15 billion parameters or exclusively examined black-box models like GPT-3 or PaLM, making replication challenging.
In the present study, we decompose the entire prompt into four components: task description, demonstration inputs, labels, and inline instructions provided for each demonstration. We investigate the effects of structural and semantic corruptions of these elements on model performance. We study models ranging from 1.5B to 70B in size, using ten datasets covering classification and generation tasks. We find that repeating text within the prompt boosts model performance, and bigger models ($\geq$30B) are more sensitive to the semantics of the prompt. Finally, we observe that adding task and inline instructions to the demonstrations enhances model performance even when the instructions are semantically corrupted. The code is available at this \href{https://github.com/text-machine-lab/Understanding_prompts_via_corruption}{URL}.
 \\ \newline  }

\begin{document}

\maketitleabstract

\section{Introduction}
\label{sec:introduction}
The ability of language models to respond to prompts and learn in context has led to an explosive growth in their use, culminating in the proliferation of AI assistants such as ChatGPT \citep{OpenAI2023GPT4TR}, Claude \citep{Claude2ModelCard}, and Bard \citep{Bard}, which use large pre-trained language models as the backbone. AI assistants built on top of backbone models are robust to prompt variation, in large part due to alignment techniques involving learning from human feedback~\citep{Ouyang2022TrainingLM}.
However, the underlying backbone models are notoriously brittle in this respect, and their performance often varies widely with slight prompt modifications. 
Building a high-quality backbone model remains a core challenge, and one of the more common ways to gauge their quality is to conduct in-context evaluation, which suffers from high sensitivity to prompt variation.
Despite this sensitivity, models have shown remarkable resilience to corruption in certain parts of the prompt.
Recently proposed explanations for in-context learning, such as implicit gradient descent \citep{dai2022can,von2022transformers}, fail to account for this resiliency.

A number of previous studies have examined the impact of prompts on model performance across different tasks
~\citep{brown2020language,radford2019language,lu2021fantastically,lialin2022life,talmor2020olmpics,webson2021prompt,lampinen,reynolds2021prompt,min2022rethinking,zhao2021calibrate,raman2022transforming, Kim2022GroundTruthLM}. However, the results have sometimes been contradictory.
In particular, the studies of individual prompt components
have been plagued by inconsistency.
For instance, \citet{webson2021prompt} found that meaningless instructions don't have a significant effect on model performance. On the other hand, evidence from \citet{mishra2021natural} and \citet{reynolds2021prompt} suggested that meaningful prompts are crucial for zero-shot performance.
Similarly, while \citet{min2022rethinking} and \citet{wei2023larger} demonstrated that label semantics aren't necessary for zero-shot performance, both \citet{Kim2022GroundTruthLM} and \citet{webson2021prompt} argued otherwise.

Additionally, the majority of prior research has focused either on smaller models with $<$15B parameters or black-box LLMs like GPT-3 \citep{brown2020language}, InstructGPT \citep{Ouyang2022TrainingLM}, and PaLM \citep{chowdhery2022palm}, and therefore don't offer a complete understanding of the significance of different prompt components across model sizes.

\begin{figure*}
    \centering
    \includegraphics[scale=0.75]{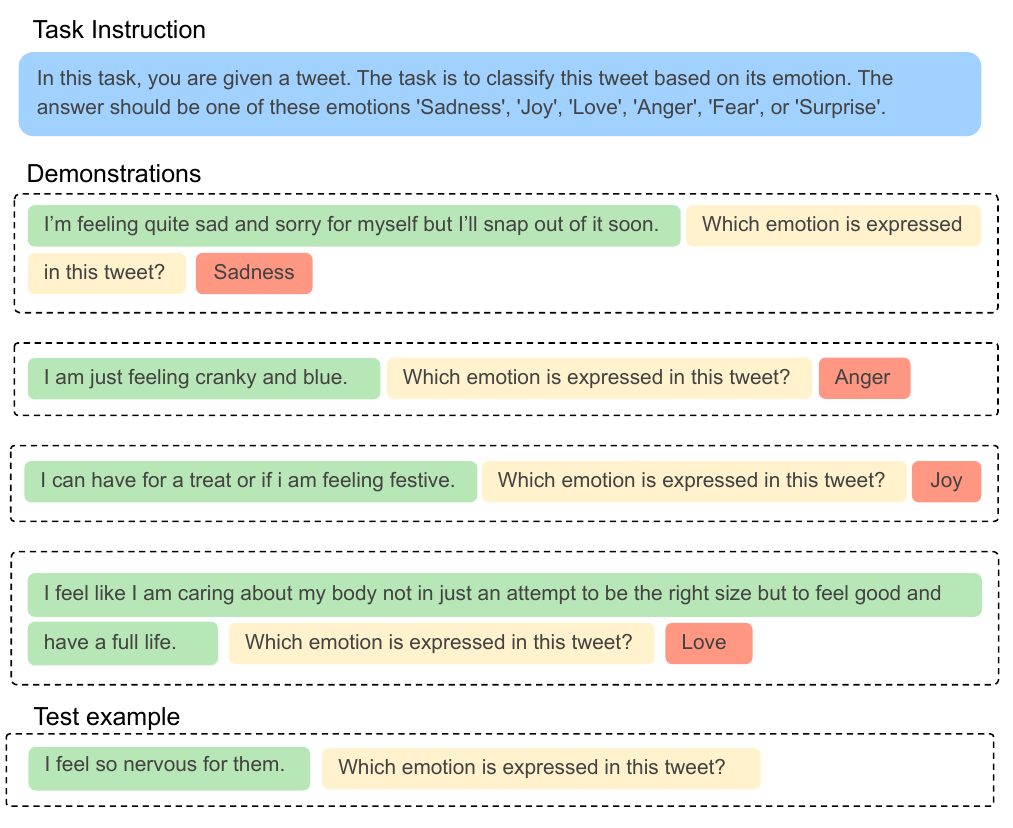}
    \caption{Prompt Components of Twitter Emotion Classification baseline prompt. Demonstration includes \colorbox{green!30}{input}, \colorbox{yellow!30}{inline instruction}, \colorbox{red!40}{label}. Two newlines are added as separators after task instruction and each demonstration. Prompts taken verbatim from Super-NaturalInstructions and PromptSource.}
    \label{fig:prompt-schema}
\end{figure*}

In the present study, we decompose the input prompt into four components: task instructions, inline instructions, and demonstrations that consist of input/target label pairs (see Figure \ref{fig:prompt-schema}) and investigate the effect of structural and semantic corruptions to these prompt components across ten models, ranging from 1.5B to 70B. We evaluate them on ten datasets, covering both classification and generation tasks. 
Building on techniques from model interpretability research, we also examine the average per component attention of two of the models to determine which components contribute more to model output. 
Our results show that:
\begin{enumerate}
    \item Including \textbf{repeated text in the prompt boosts model performance}.
    \item Addition of both \textbf{task and inline instructions improves model performance}, even when these instructions are random words.
    \item \textbf{Larger models exhibit higher sensitivity to prompt semantics} and pay more attention to the semantically relevant prompt components.
\end{enumerate}


\section{Related work}
\label{sec:related-work} 
Several prior studies have investigated in-context learning (ICL)  capabilities of large language models
~\citep{brown2020language,radford2019language,lu2021fantastically,lialin2022life,talmor2020olmpics,webson2021prompt,lampinen,reynolds2021prompt,min2022rethinking,zhao2021calibrate,raman2022transforming,wei2023larger,Madaan2022TextAP}. 
However, when it comes to the impact of different parts of the prompt on model performance, the conclusions have often been inconsistent.
For example, ~\citet{webson2021prompt} suggest that relevant and irrelevant instructions in the prompt yield similar model performance, whereas \citet{mishra2021natural} and \citet{reynolds2021prompt} argued the opposite. The latter studies showed that detailed and task-relevant prompts that closely resemble natural human language give better model performance.
Similarly, \citet{Kim2022GroundTruthLM} studied the importance of ground-truth labels for in-context learning and found that ground-truth labels were important for ICL, contradicting the results from   \citet{min2022rethinking}.

While prior work has not provided a comprehensive analysis of the impact of different prompt components on model performance, a few studies have selectively examined specific elements of the prompt.
For example, 
~\citet{lampinen} looked into adding explanations to the demonstration and found that adding task explanations can significantly improve model performance.
\citet{min2022rethinking} examined different aspects of in-context demonstrations and found that input-label mapping did not significantly affect model accuracy.
\citet{webson2021prompt} and \citet{gu2021ppt} studied instructions and labels and suggested that the labels were more important than instructions. 
\citet{wei2023larger} investigated the effect of semantic priors associated with the labels during pre-training, relative to the input-label mapping provided in the prompt, showing that the ability to override semantic priors with the prompt is an emergent ability. 

Prior work has also examined different prompting strategies, as well as additional fine-tuning to improve in-context performance. 
For instance, \citet{xu2023re} introduced re-reading prompting strategy where they repeat the question in the prompt and found that this strategy improves performance for ChatGPT and GPT-3. 
\citet{wei2023symbol} proposed ``symbol tuning'', fine-tuning models with arbitrary labels, and observed performance improvements on unseen ICL tasks.
In a different approach, \citet{gonen2022perplexity} proposed constructing the prompt with lower perplexity for better performance. 

Few studies \cite{dai2022can,von2022transformers} also linked attention computation performed during in-context learning to model updates performed with gradient descent. However, it is unclear how this mechanism would account for some aspects of in-context learning, such as the success of zero-shot prompting.

\section{Experiment setup}
\subsection{Prompts}

\paragraph{Prompt components}
We use the term ``\textbf{prompt}'' to refer to the complete input text provided to the model. A prompt consists of four main components: a task instruction, demonstration input, demonstration label, and brief inline instructions accompanying each demonstration (see Figure \ref{fig:prompt-schema}).
Two newlines are used as a separator after the task instruction and after each demonstration.
We refer to a prompt with all components, including a test instance and its inline instruction, as a \textbf{baseline prompt}.
Our experiments are conducted in a zero-shot and 4-shot setting. Figure \ref{fig:prompt-schema} shows the baseline prompt for Twitter Emotion classification dataset. Baseline prompts for all the datasets are provided in the Appendix \ref{baselines_prompts_for_all_datasets}.

\paragraph{Prompt design}
We leverage the task instructions and demonstrations provided by \citet{wang2022super} for each dataset, as they have been reviewed and refined through multiple iterations. We use inline instructions from PromptSource \cite{Bach2022PromptSourceAI}. 
To ensure coherence and simplicity, we made a few changes to the task and inline instructions, following the recommendations of \citet{gonen2022perplexity}.

\paragraph{Prompt corruptions}
We perform two types of prompt corruptions: \textbf{structural corruption} and \textbf{semantic corruption}. In structural corruption, we add or remove the prompt components, depending on the setup. We start with the test instance and add components one by one to analyze their effect on model performance. 
To assess the impact of repeated text in the prompt, we systematically eliminate inline instructions from the baseline prompt. We remove the inline instruction from one demonstration, then two, and continue this process until we have removed the inline instructions from all four demonstrations. These corruptions are referred to as \textbf{repeated text corruptions}. We keep the inline instruction which follows the test instance as is. 

In semantic corruption, we disrupt the semantics of prompt components.  Task and inline instructions are corrupted with random words drawn from the \texttt{english\_words}\footnote{https://pypi.org/project/english-words/} set. With a 100\% corruption rate, we refer to this corruption as the \textbf{random words} corruption.
The random word instructions retain the same number of tokens as the original (meaningful) instructions. Labels are perturbed by assigning incorrect labels to the demonstrations. These incorrect labels are drawn from the same label space. This corruption is referred to as the  \textbf{wrong label} corruption and is only applied to classification tasks.  
In the \textbf{random words label} corruption, we replace original labels with random words, similar to the instruction random words corruption, but we use the original labels to assess the model's predictions.
To noise the demonstration inputs, we replace them with random sentences sampled from Common Crawl. We refer to this as \textbf{Out-Of-Distribution (OOD)} input corruption.

\subsection{Models, datasets and metrics}
\paragraph{Models}
To cover a broad range of models, we conducted experiments with ten models ranging in size from 1.5B to 70B. The models are GPT2-xl \cite{radford2019language}, GPT-J-6B \cite{wang2021gpt}, Pythia-12B \cite{Biderman2023PythiaAS}, OPT-30B, OPT-30B-IML-MAX\footnote{Instruction tuned variant of OPT. 
}, OPT-66B \cite{zhang2022opt}, Vicuna-33B \cite{vicuna2023}, Llama-7B, Llama-2-70B and Llama-2-70B-chat \cite{Touvron2023Llama2O}. This provides a wide range of model sizes, and types and also doesn't focus on a single model family, making the results more generalizable. Our study included pre-trained language models, as well as instruction-tuned and aligned models. We refer to models as "aligned" when they undergo additional training through reinforcement learning from human feedback (RLHF) \cite{Ouyang2022TrainingLM}.

\paragraph{Datasets}
The evaluation was conducted on ten datasets from Super-NaturalInstructions \cite{wang2022super}.  Datasets include eight classification tasks: RTE, Medical Question Pair, Financial Phrasebank, Twitter Emotion classification, CoLA, AgNews, COPA, Com2sense, and two generation tasks: TriviaQA and Mathdataset answer generation \cite{wang2022super}. Following \citet{wang2022super}, we used 100 randomly sampled balanced test instances for each of the 10 datasets. Data statistics are shown in Table \ref{tab:data_statistics} in the Appendix.

\paragraph{Evaluation method and metrics}
 For evaluation, we used Exact Match for classification tasks, and Rouge-L for generative tasks. Following \citet{wang2022super}, we strip the model response at the first full stop symbol. We used the jackknife variance estimate method to calculate the mean 
 of model performance. The mean performance, averaged across tasks, is reported for each model in Tables \ref{tab:avg_result_across_dataset_4models} and \ref{tab:avg_result_across_dataset_10models}. For all figures, we plot the mean as the ``average score''. For generation tasks, we used the greedy decoding strategy, setting the top-p and temperature values to 1 and limiting the maximum number of new tokens to 10.

\paragraph{Attention computation}
To understand the significance of different prompt components, we computed the average attention norm per prompt component for GPT-J-6B \cite{wang2021gpt} and OPT-30B \cite{zhang2022opt}. 
Following \citet{kobayashi2020attention}, we compute the L2 norm of the sum of the attention-weighted value vector $\| \sum \alpha V(x) \|$, where $\alpha$ is the attention weight, $x$ is the input vector, and $V(x) = {W_O}(W_Vx)$ is the value projection of $x$, followed output transformation $W_O$.
Specifically, we used the last token of the prompt as the query token and extracted attention norms for the other tokens. For each token, we averaged these norms across all layers. We then averaged the resulting scores over all tokens corresponding to a given prompt component. This average is reported in Figures \ref{fig:bothmodel_attn}, \ref{fig:gptj_attn} and \ref{fig:opt30b_attn}.

%

%
 For GPT-J-6B, each plot shows the average attention norm over 100 samples (10 samples per dataset). 
For OPT-30B, due to computing costs, we focused on datasets with shorter baseline prompts: CoLA, Twitter Emotion classification, and TriviaQA. For each attention plot, we included the average attention norm from 30 samples, selecting only those where the model predictions were correct.

\section{Results}
\label{result}

Tables \ref{tab:avg_result_across_dataset_4models} and \ref{tab:avg_result_across_dataset_10models} show results for each corruption across 10 datasets. Here's a breakdown of the prompt configurations used:
\begin{itemize} 
 \setlength{\itemsep}{-0.5em}
    \item \textbf{Test instance}: The input prompt containing only the test instance.
    \item \textbf{+task instr.}: Test instance with the task instruction added.
    \item \textbf{+inline instr.}: Test instance with an inline instruction added instead of the task instruction.
    \item \textbf{+both instr.}: Test instance with both task and inline instructions added.
    \item \textbf{+demos.}: Test instance plus four demonstrations (no instructions included).
    \item \textbf{+task instr. +demos.}: Test instance with task instructions and four demonstrations.
    \item \textbf{+inline instr. +demos.}: Test instance with four demonstrations (each containing an inline instruction) and no task instruction.
    \item \textbf{Baseline}: Includes all components (task instruction, inline instruction, demonstrations, and test instance).
    \item \textbf{Baseline -inputs}: Baseline prompt with demonstration inputs removed.
    \item \textbf{Baseline -labels}: Baseline prompt with demonstration labels removed.
    \item \textbf{Rw both instr.}: Random word corruption applied to both task and inline instructions.
    \item \textbf{Rw labels}: Random word corruption applied to labels.
    \item \textbf{OOD inputs}: Out-of-distribution input corruption.
    \item \textbf{Inline instr. in [n] demos.}: Meaningful inline instructions added to "n" demonstrations.
    \item \textbf{Rw inline instr. in [n] demos.}: Random word inline instructions added to "n" demonstrations.
\end{itemize}

\begin{figure*}[h]
    \centering
    \includegraphics[scale=0.7]{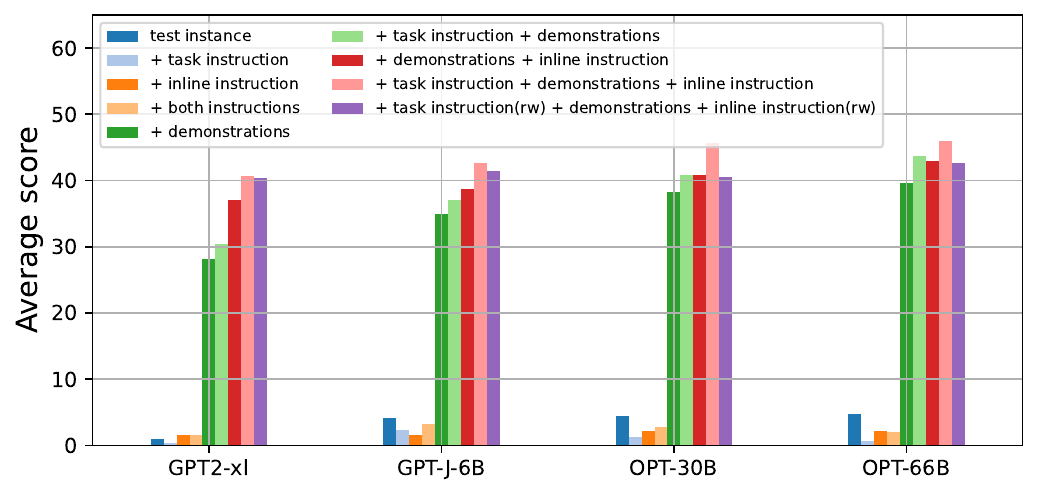}
    \caption{ Demonstrations improve the average score, adding task and inline instruction improves it further, even when instructions are just random words.  The Y-axis represents the average score across all datasets. The use of random words is indicated with ``rw''.}
    \label{fig:4_models_effect_of_instruction_1a}
\end{figure*}
\begin{figure*}[h]
    \centering
    \includegraphics[scale=0.6]{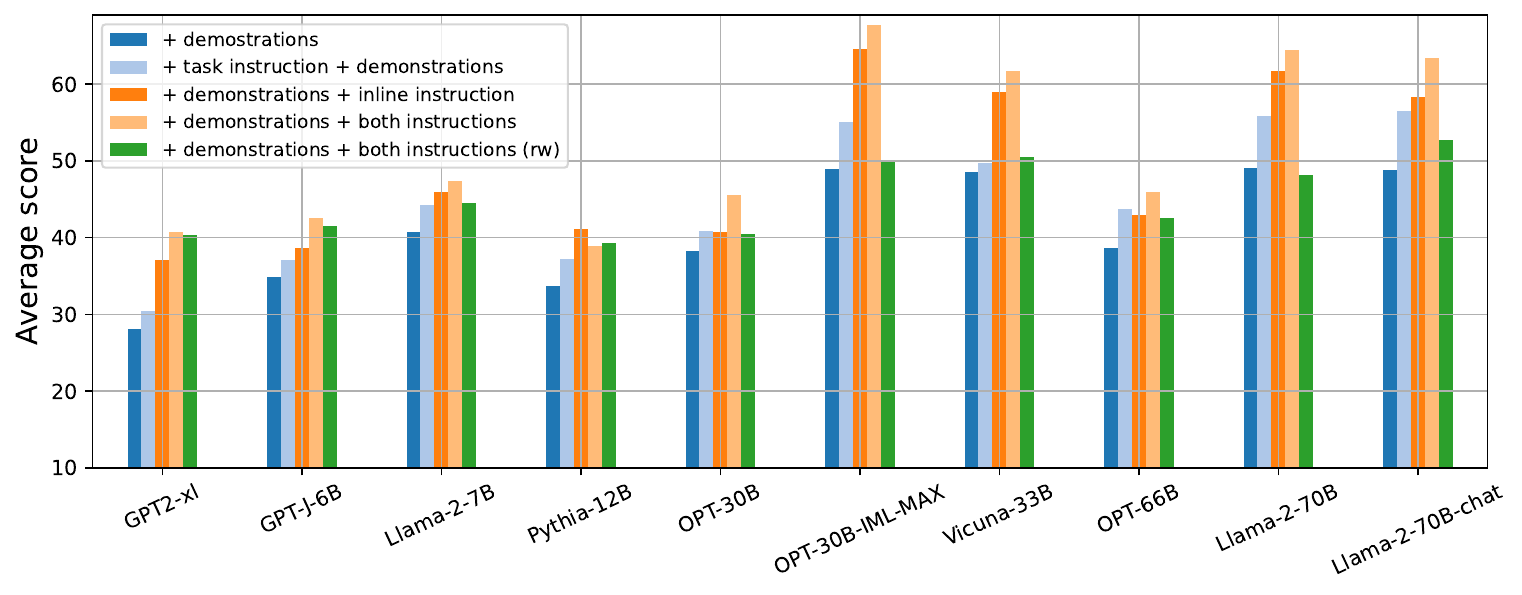}
    \caption{ Adding relevant or meaningless instruction to the prompt improves model performance. The components are added to the test instance. For example `+ demonstrations' means test instance + demonstration. The Y-axis represents the average score across all datasets. Random words are indicated with ``rw''.}
    \label{fig:1b}
\end{figure*}

\paragraph{Adding task and inline instructions boosts the performance even when the instructions are random words.}
Our experiments with four models (see Figure \ref{fig:4_models_effect_of_instruction_1a}) highlight that the addition of demonstrations to the test instance has the most impact, producing a gain of 25-35\% across all models. Accuracy is improved further by adding task instructions and inline instructions (Figure \ref{fig:4_models_effect_of_instruction_1a}). The models gain between 5-18\% accuracy when meaningful instructions are added. Interestingly, this gain is between 1-12\% when instructions are just random words, except for Llama-70B. Figure \ref{fig:1b} shows a similar effect for all ten models.  

\begingroup
\begin{table*}
    \centering
    \begin{tabular}{ p{0.45\columnwidth} | c c c c}
    \toprule
      Structural Corruptions & GPT2-xl & GPT-J-6B & OPT-30B & OPT-66B \\
        \midrule   
        Test instance &  0.9 & 4.2 & 4.4 & 4.7 \\ 
        + task instr. & 0.4 & 2.3 & 1.2 & 0.7 \\ 
            + inline instr. & 1.6 & 1.6 & 2.2 & 2.1 \\ 
        + both instr.  &  1.6 & 3.3 & 2.7 & 2.0 \\ 
        + demo.  &  28.1 & 34.9 & 38.3 & 39.6 \\ 
        + task instr. + demo.  &  30.4 & 37.1 & 40.9 & 43.7 \\  
        + inline instr + demo.  &  37.1 & 38.7 & 40.8 & 43.0 \\ 
        Baseline  & \underline{\textbf{40.7}} & \underline{\textbf{42.6}} & \underline{\textbf{45.6}} & \underline{\textbf{45.9}} \\
        Baseline - labels & 0.1 & 0.2 & 0.5 & 0.9 \\ 
        Baseline - inputs &  27.1 & 17.0 & 22.0 & 22.3 \\         
        \bottomrule
    \end{tabular}
\caption{Model performance averaged across all datasets. The highest performance is in bold, \textit{baseline prompt} performance is underlined.}
\label{tab:avg_result_across_dataset_4models}
\end{table*}
\endgroup

\begin{table*}
\centering
\begin{adjustbox}{width=\textwidth}
\begin{tabular}{l*{10}{r}}
\toprule
\textbf{Corruptions} & \makecell{GPT2\\xl} & \makecell{GPT-J\\6B}  & \makecell{LLama\\7B}  & \makecell{Pythia\\12B} & \makecell{OPT\\30B} & \makecell{OPT-30B\\IML-MAX}  & \makecell{Vicuna\\33B}  & \makecell{OPT\\66B} & \makecell{LLama-2\\70B}  & \makecell{LLama-2\\70B-chat} \\

\midrule
\makecell[l]{\textbf{Structural}} \\
+ demos. & 28.1 & 34.9 & 40.8 & 33.7 & 38.3 & 49.0 & 48.5 & 38.6 & 49.1 & 48.8 \\
+ task instr. + demos. & 30.4 & 37.1 & 44.2 & 37.2 & 40.9 & 55.1 & 49.7 & \textbf{43.7} & 55.9 & 56.5 \\
+ inline instr. + demos. & 37.1 & 38.7 & 46.0 & \textbf{41.1} & 40.8 & 64.6 & 59.0 & 43.0 & 61.7 & 58.3 \\
Baseline & \underline{40.7} & \underline{42.6} & \underline{\textbf{47.4}} & \underline{38.9} & \underline{\textbf{45.6}} & \underline{\textbf{67.7}} & \underline{\textbf{61.7}} & \underline{\textbf{45.9}} & \underline{\textbf{64.5}} & \underline{\textbf{63.4}} \\
\midrule
\makecell[l]{\textbf{Semantic }} \\
Rw both instr. & 40.4 & 41.5 & 44.5 & 39.3 & 40.5 & 49.8 & 50.5 & 42.6 & 48.2 & 52.7 \\
Rw labels & 3.7 & 1.8 & 1.4 & 1.4 & 3.4 & 46.5 & 3.8 & 2.7 & 1.2 & 7.9 \\
OOD inputs & \textbf{41.7} & 40.7 & 43.9 & 38.1 & 44.6 & \textbf{67.6} & 57.1 & 40.7 & 50.5 & 57.4 \\
\midrule
\makecell[l]{\textbf{Repeated Text }}\\
Inline instr. in 3 demos. & \textbf{43.2} & \textbf{43.2} & \textbf{48.2} & 39.3 & \textbf{45.8} & 65.8 & \textbf{61.1} & 43.1 & \textbf{64.6} & \textbf{63.5} \\
Inline instr. in 2 demos. & 40.9 & \textbf{43.1} & 44.5 & \textbf{41.6} & 43.7 & 63.9 & 59.3 & 43.5 & 62.7 & 62.6 \\
Inline in instr. 1 demos. & 40.6 & \textbf{43.1} & 42.7 & 39.8 & 45.7 & 63.2 & 58.8 & 41.9 & 62.0 & 61.3 \\
Inline in instr. 0 demos. & 13.3 & 22.6 & 17.9 & 14.8 & 21.4 & 59.0 & 30.5 & 20.3 & 35.1 & 29.2 \\
Rw inline instr. in 3 demos. & 35.8 & 38.8 & 43.1 & 38.8 & 35.6 & 50.9 & 51.9 & 39.7 & 48.2 & 56.1 \\
Rw inline instr. in 2 demos. & 35.9 & 36.1 & 41.1 & 36.0 & 35.4 & 45.4 & 46.6 & 40.6 & 49.2 & 49.0 \\
Rw inline instr. in 1 demos. & 33.0 & 34.9 & 36.7 & 31.4 & 22.4 & 35.0 & 38.7 & 32.8 & 37.7 & 41.5 \\
Rw inline instr. in 0 demos. & 0.6 & 0.2 & 1.3 & 0.4 & 0.2 & 0.2 & 1.1 & 0.3 & 0.7 & 1.6 \\
\bottomrule
\end{tabular}
\end{adjustbox}
\caption{Model performance averaged across all datasets. Structural corruption is when components are added to the test instance. Repeated text corruptions are performed on baseline prompt which includes inline instruction in all four demonstrations. Random words text is represented by ``Rw''. The top two performances for each model are in bold, and baseline prompt performance is underlined.
}
\label{tab:avg_result_across_dataset_10models}
\end{table*}

\paragraph{Inline instructions are more important than task instructions.}
From Figure \ref{fig:1b}, we see that the performance gained by inline instruction is 2.5-12.5\% across models whereas task instruction helps models by only 1-7.5\%. This pattern is observed for models of all sizes, except OPT-66B, where the gain obtained by the inline instructions is close to that of the task description (Figure \ref{fig:1b}). This suggests models benefit from the brief repetitive text more than from a detailed task instruction.

\begin{figure*}
    \centering
    \includegraphics[scale=0.6]{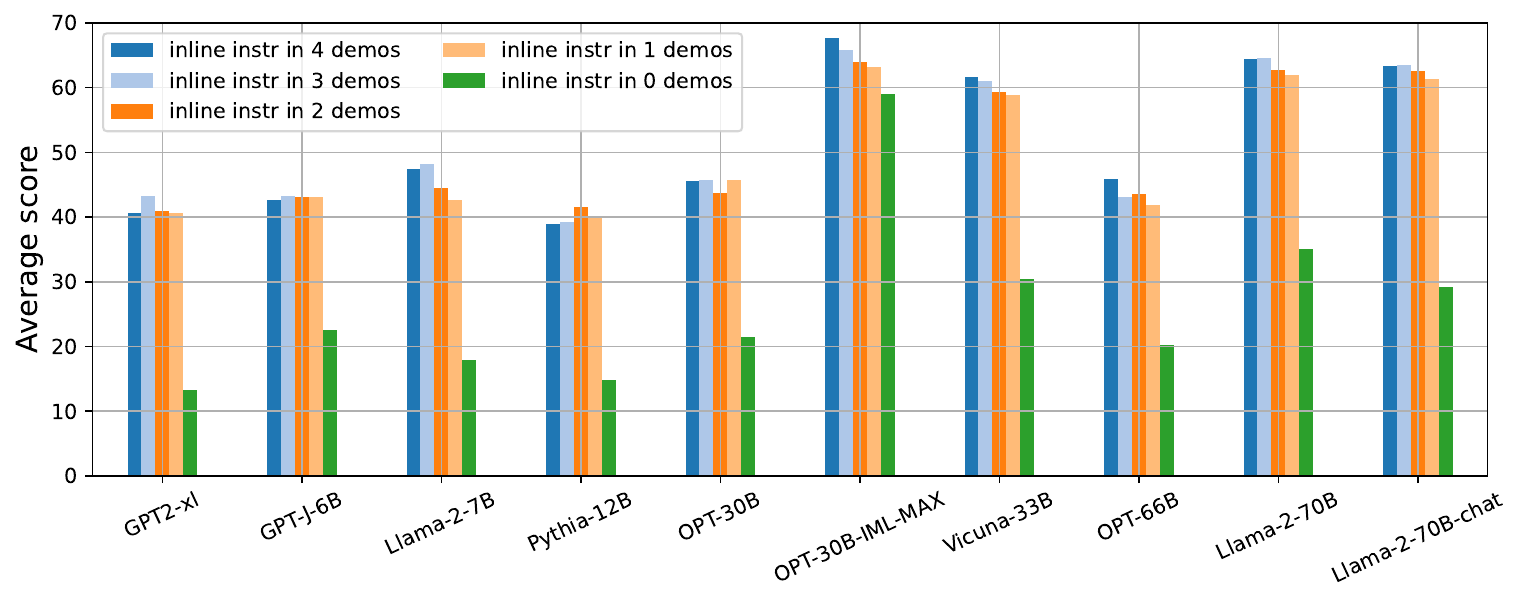}
    \caption{ Repeated text boosts performance. Inline instruction in four demos is the baseline prompt. Inline instruction which occurs after the test instance is kept as is.}
    \label{fig:4b}
\end{figure*}

\begin{figure*}
    \centering
    \includegraphics[scale=0.58]{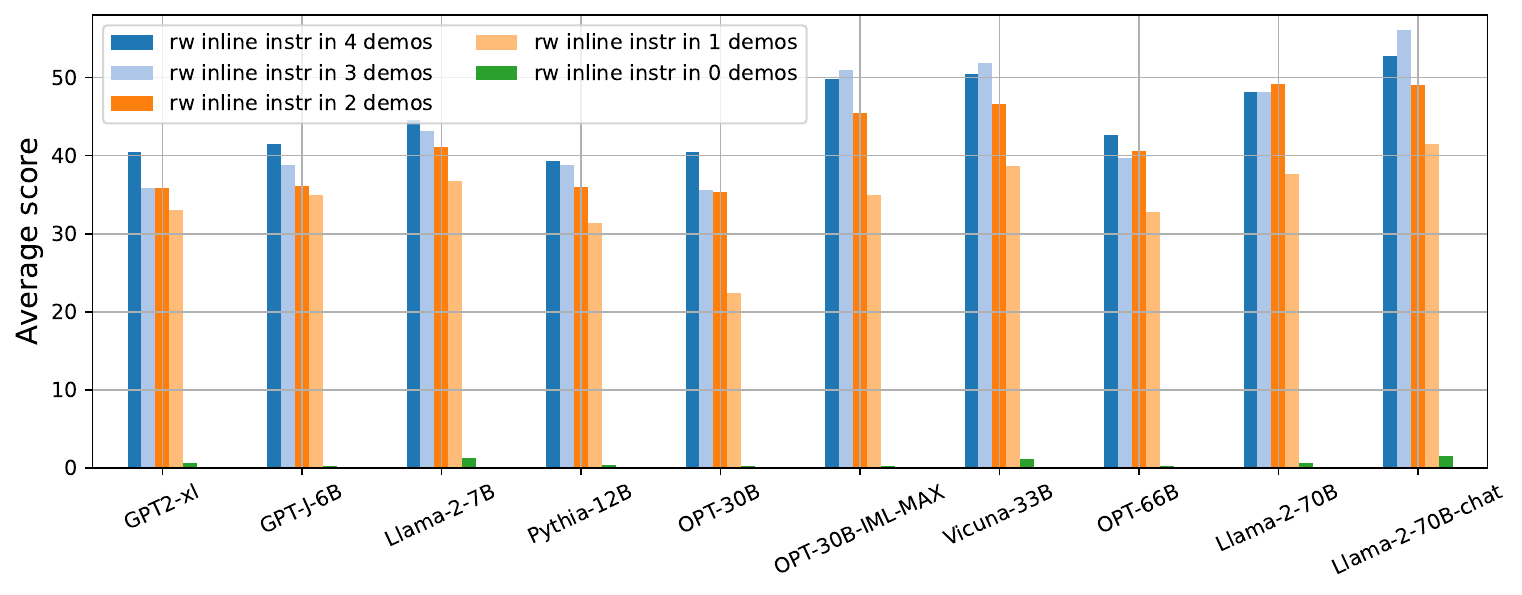}
    \caption{Repeated text boosts performance even when the text is irrelevant; ``rw'' refers to random words. The prompts include all components but the instructions are replaced with random words.}
    \label{fig:5b}
\end{figure*}

\paragraph{Repeated text boosts performance.}
We further investigated the effects of repeating inline instructions. In Figure \ref{fig:4b}, we plot the results for the baseline prompt (which includes all components) and the results obtained when eliminating inline instructions from demonstrations one by one. Note that we always keep the inline instruction that occurs after the test instance. We see a huge drop in performance when removing the inline instruction from all demonstrations. The drop is 20-35\% across all models, except OPT-30B-IML, which shows a drop of 8.8\%. Interestingly, we observed a similar effect for prompts in which inline instructions were replaced with random words, producing a performance drop of 40-51\% (cf. Figure \ref{fig:5b}). This suggests that the mere presence of repetitive text in the prompt, whether relevant or irrelevant, can boost model performance.

\begin{figure*}
    \centering
    \includegraphics[scale=0.55]{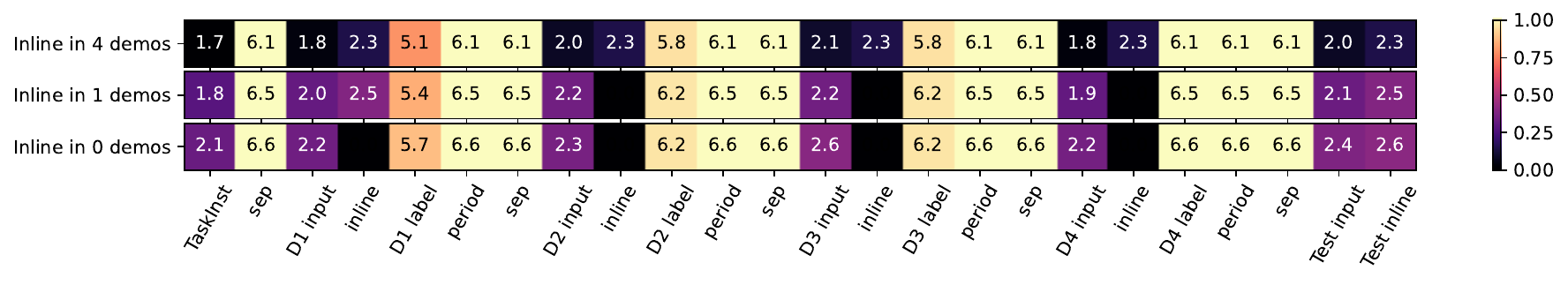}
    \caption{ Average OPT-30B attention per component for repeated text corruptions. ``Inline'' refers to the presence of the number of inline instructions in the baseline prompt. A solid black box represents omitted  components. }
    \label{fig:opt-30b_repeatedtextcorrup_attention}
\end{figure*}

However, how often we introduce these repetitions in the prompt also matters. Table \ref{tab:avg_result_across_dataset_10models} shows that models like OPT-30B and Llama-70B can achieve better performance with only two or three meaningful inline instructions in the prompt.
The attention plot in Figure \ref{fig:opt-30b_repeatedtextcorrup_attention} shows that if we introduce inline instruction in four demonstrations rather than in one, the attention to the input segment of each demonstration dropped from 2\% to 1.8\% and the attention to each of the labels decreased by around ~0.4\%.
%
We see a similar pattern for repeated text corruptions in prompts with random word instructions (see Figure \ref{fig:opt-30b_repeatedtextcorrup_attention_rw} in Appendix).

\begin{figure*}
    \centering
    \includegraphics[scale=0.6]{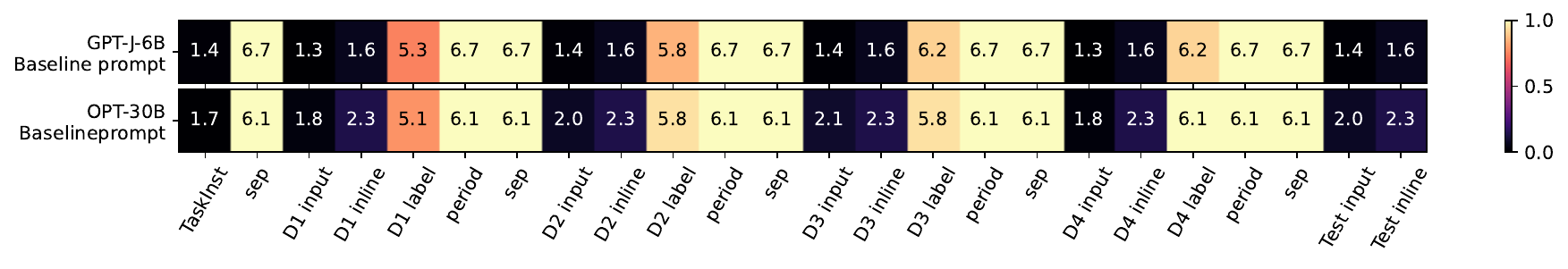}
    \caption{
    Average attention per component for GPT-J-6B and OPT-30B baseline prompts. `D' stands for Demonstration and ``sep'' for the new line separator.
    }
    \label{fig:bothmodel_attn}
\end{figure*}

\begin{figure}
    \centering
    \includegraphics[scale=0.6]{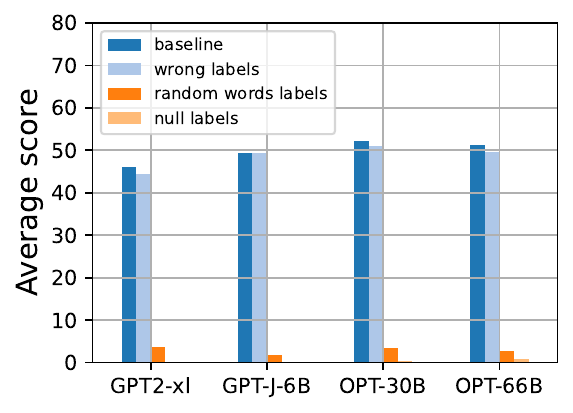}
    \caption{Label from label space is important. Complete removal of labels drops the performance to almost zero. }
    \label{fig:2a_labels_4_models}
\end{figure}

\begin{figure*}
    \centering
    \includegraphics[scale=0.6]{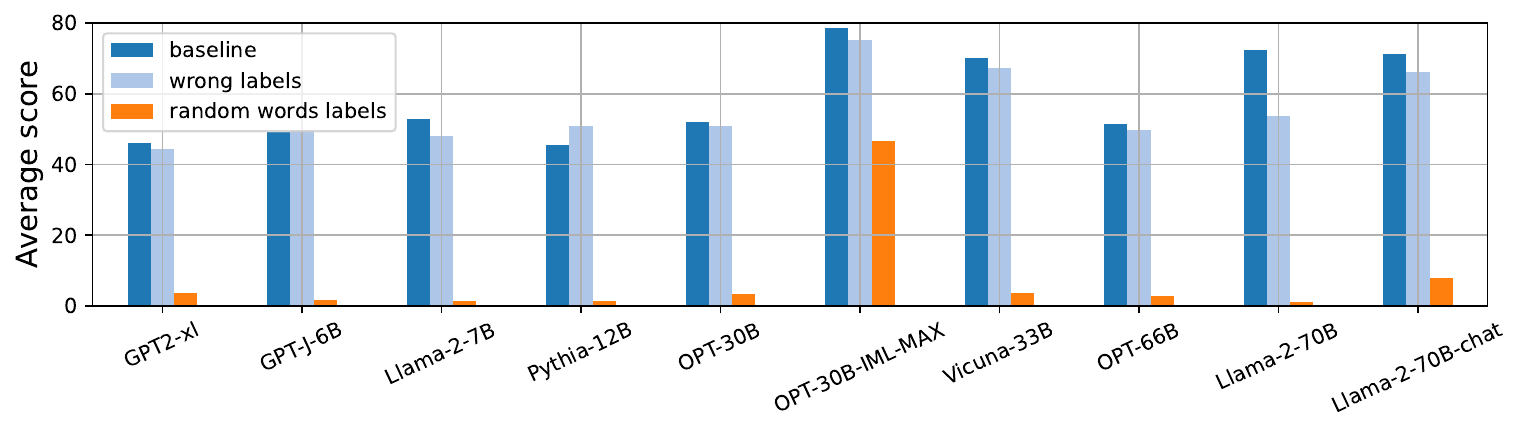}
    \caption{ Using labels from the correct label space is crucial for model performance.}
    \label{fig:2b_labels_all_models}
\end{figure*}

\paragraph{Labels must be drawn from the label space, but need not be correct.}
When we perturb labels with the \textbf{wrong label} corruption, the performance drops just by 0-6\% across all models (except for Llama-2-70B, where the drop is 18.6\%). However, when we apply \textbf{random words label} corruption, the accuracy drops to almost zero for all models (except OPT-30B-IML-MAX) (cf. Figure \ref{fig:2b_labels_all_models}). Complete removal of the labels from the prompt has a similar effect (cf. Figure \ref{fig:2a_labels_4_models}). 



\begin{figure}
    \centering
    \includegraphics[scale=0.5]{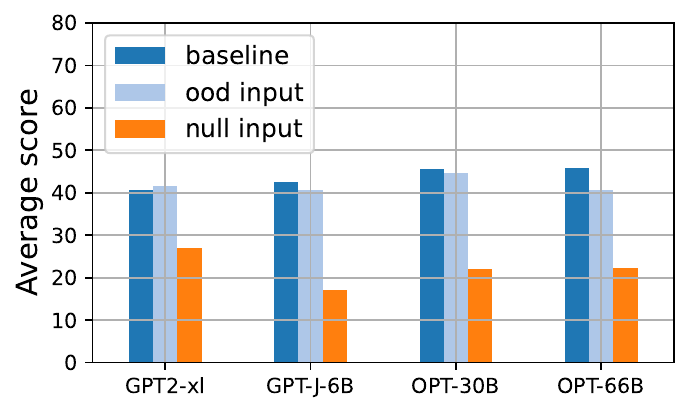}
    \caption{ Semantics of the demonstration input is not important. Complete removal of labels drops the performance.}
    \label{fig:3a_inputs_4_models}
\end{figure}

\begin{figure*}
    \centering
    \includegraphics[scale=0.6]{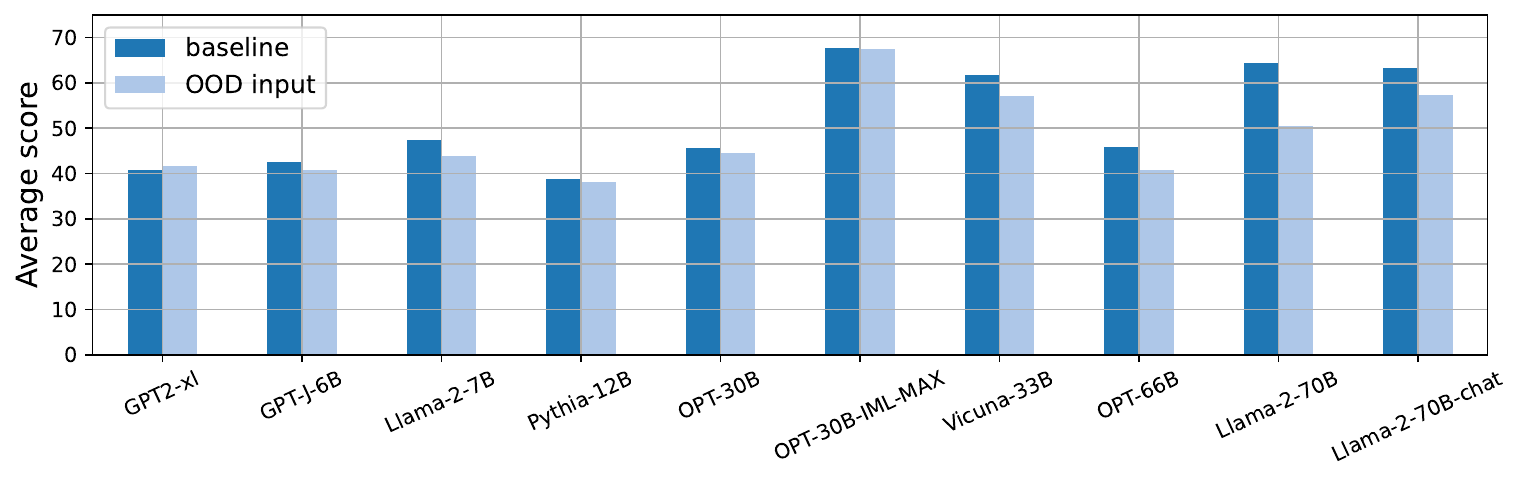}
    \caption{Semantics of the demonstration input is not important.}
    \label{fig:3b_inputs_all_models}
\end{figure*}


\paragraph{Bigger models are more sensitive to the semantics of the prompt.}
We divide the models in the study into smaller ($<$15B) and bigger ($\geq$ 30B) models. In Figure \ref{fig:1b}, smaller models show the performance gain between 5-12\% with both relevant and irrelevant instructions, whereas bigger models gain more with meaningful instructions (7-18\%) and just 1-4\% with random word instructions. When we perturb demonstration inputs with OOD sentences (see Figure \ref{fig:3b_inputs_all_models}), smaller models' accuracy drops by 1-4\%. In bigger models, this performance decrease is larger (1-6\%), with Llama-2-70b showing a huge drop of 18\%, which suggests that bigger models are more sensitive to prompt semantics. 

\begin{figure*}
    \centering
    \includegraphics[scale=0.55]{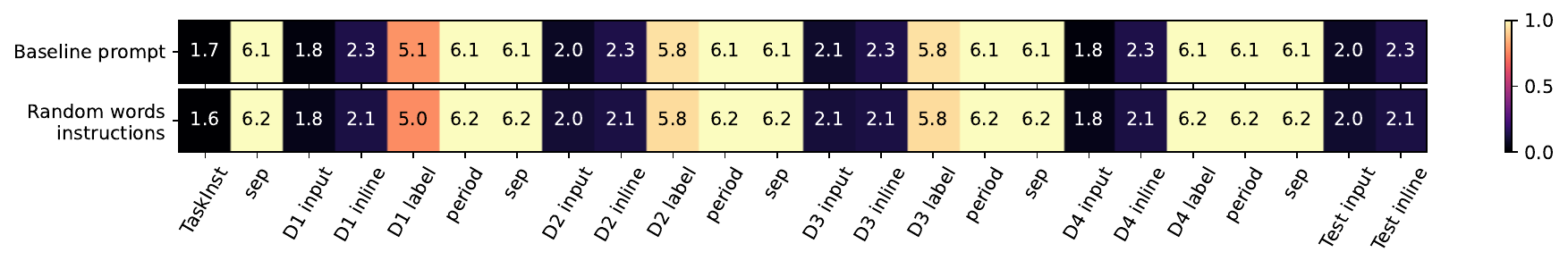}
    \caption{ Average attention per component for OPT-30B: Baseline prompt versus prompt when both task and inline instructions are replaced by random words.}
    \label{fig:opt30b_attn}
\end{figure*}

\begin{figure*}
    \centering
    \includegraphics[scale=0.55]{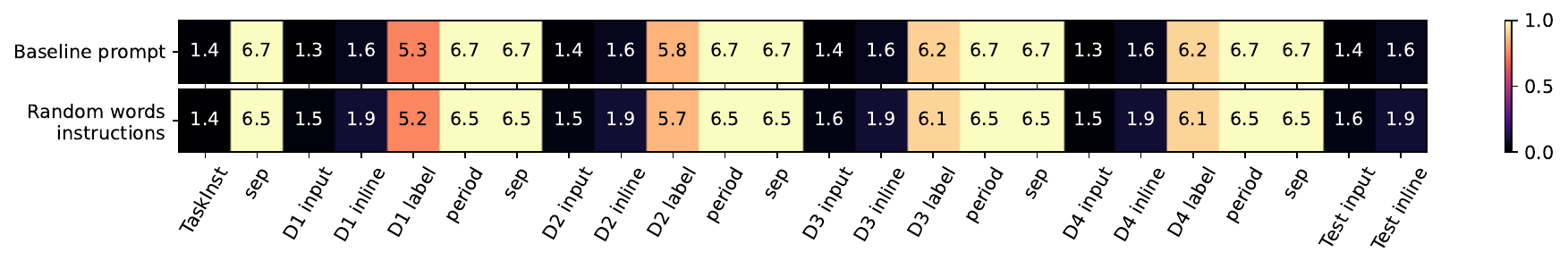}
    \caption{ Average attention per component for GPT-J-6B. Baseline prompt versus prompt when both task and inline instructions are replaced by random words.}
    \label{fig:gptj_attn}
\end{figure*}

\paragraph{Bigger models pay more attention to relevant components.}
In Figure \ref{fig:bothmodel_attn}, we plot the average attention per component for GPT-J-6B and OPT-30B baseline prompts. 
In line with earlier observations about vertical attention patterns  \cite{kovaleva2019revealing}, we find that the models allocate the highest attention weight to separators, after which the most attended to component is labels. Inline instructions are next, followed by demonstration inputs and task instructions. 
%
Compared to smaller models, larger models seem to allocate more attention to relevant components and less to separators when generating the target label. For example, OPT-30B allocates 5.4\% less attention to separators compared to GPT-J-6B, and instead increases attention to inline instructions by 3.5\% and to demonstration inputs by 2.3\%.
%
To compare how models behave when text is corrupted semantically, we plotted attention for prompts with meaningful versus irrelevant instructions (see Figures \ref{fig:opt30b_attn} and \ref{fig:gptj_attn}). We see that GPT-J-6B shifts its attention from separators and labels to the demonstration input and the random inline instructions. OPT-30B does the opposite: it reduces its attention to random words text and shifts it to the separator.
%


\paragraph{Results are similar in classification and generation tasks, with few exceptions.}
As can be seen in Table \ref{tab:avg_result_across_class_n_gen_tasks}, 
repeating inline instructions has a big impact on both classification and generation tasks regardless of model size. The first repetition has the most pronounced effect, and this is true even when the inline instructions are random. To see this effect, compare the rows ``Inline instr. in 0 demos'' and ``Inline instr. in 1 demos'', as well as the rows ``Rw Inline instr. in 0 demos'' and ``Rw Inline instr. in 1 demos'' in Table \ref{tab:avg_result_across_class_n_gen_tasks}. 
We also observe that adding relevant or random word instructions to demonstrations improves GPT2-xl performance by an average of 12.5\% for classification and 3.8\% for generation tasks. For the larger model LLama-2-70B, relevant instructions lead to gains of 18.5\% for classification and 3.3\% for generation tasks. Adding random word instructions yields LLama-2-70B, a marginal improvement of 0.1\% in classification tasks and a drop of 1.7\% for generation tasks. To see this effect, compare the rows ``+demos.'', ``Baseline'' and ``Rw both instr.'' in Table \ref{tab:avg_result_across_class_n_gen_tasks}.  The performance drop from meaningful to random word instructions is more pronounced (20.2\% drop) in LLama-2-70B in classification tasks (compare rows ``Baseline'' and ``Rw both instr.'' in Table \ref{tab:avg_result_across_class_n_gen_tasks}) suggesting that larger models pay more attention to the meaning of the instruction. In generation tasks, both model sizes exhibit a comparable drop in performance.

\paragraph{Results are consistent across most of the datasets.}
Tables  \ref{tab:avg_result_across_dataset_tasks_gpt2-xl} and \ref{tab:avg_result_across_dataset_tasks_llama-70b} show results for each dataset individually for both GPT2-xl and LLama-2-70B.  The first repetition of relevant or irrelevant inline instructions in the prompt significantly boosts performance across all datasets for both model sizes. Adding relevant instructions proves beneficial for both GPT2-xl and LLama-2-70B on all 10 datasets. At the same time, adding random word instruction benefits GPT2-xl on 8 out of 10 datasets, but LLama-2-70B only on 6 out of 10 datasets. Corrupting labels with random words impacts GPT-2-xl on 8 out of 10 datasets and LLama-2-70B on all datasets.


\section{Conclusion}
%
This study investigated the importance of different components of a prompt for large language models. We systematically corrupted prompts in different ways across 10 models ranging from 1.5 billion to 70 billion parameters and evaluated their performance on 10 diverse datasets. We also examined how much attention the models allocate to different prompt components.  One interesting finding was that adding any inline instructions to the prompt, even just random words, actually helps models perform better.  We also showed that repeated text improves model performance drastically and that larger models are substantially more sensitive to prompt semantics. We hope our study will pave the way for more refined and effective prompting strategies in future applications.

\section{Limitations}
Our study was focused on exploring various types of corruption across a diverse range of datasets and model sizes. It involved a large number of experiments and certain prompt elements were held constant, such as demonstrations and instructions. Altering these components might introduce variations in the results, and this aspect should be taken into consideration for further research.
Additionally, we limited the attention analysis to datasets with shorter prompts due to the computational intensity and cost associated with computing attention norms.
An additional limitation arises from the use of the same prompt template across all model types. This uniformity may lead to some performance discrepancies in instruction-tuned models.

\section{Ethic Statements}
Our goal with this study is to enrich the understanding of prompting and contribute to the responsible utilization of large language models. 
We believe that attention analysis can offer meaningful insights to the research community, facilitating the development of more robust language models. 
It's noteworthy that all the models and datasets employed in our research are open source, and we meticulously reported all experiment details in the paper to support the transparency and accessibility of the research.

\section{Acknowledgement} This work was funded in part by an Amazon Alexa AI research award to Anna Rumshisky. We would like to express our gratitude to Anton Kovalev for helping with the tables in the paper.

\nocite{*}
\section{References}\label{sec:reference}

\bibliographystyle{lrec-coling2024-natbib}
\bibliography{custom}


\clearpage
\appendix

\section{Components of Prompts for all datasets}
\label{baselines_prompts_for_all_datasets}
We show baseline prompts for all datasets. We use 4-shot setting and each prompt consists of four components: \textcolor{cyan}{Task instruction}, \textcolor{orange}{inline instruction}, \textcolor{gray}{demonstration input} and \textcolor{green}{label}. [Test instance] will vary. 
Each dataset consists of 100 samples and is balanced. Data statistics for each of the datasets is shown in \ref{tab:data_statistics}.

\begingroup
\begin{table*}
    \centering
    \begin{tabular}{cc}
        \hline
        \textbf{Dataset} & \textbf{Statistics} \\
        \hline
        RTE & Yes(50), No(50)\\
        Medical Question Pair &  Similar(50), Similar(50) \\
        Financial Phrasebank & Neural(33), Negative(33), Positive(34)\\
        Twitter Emotion classification & Sadness(17), Joy(17), Love(17),
Anger(17), Fear(16), Surprise(16).\\
        CoLA & Yes(50), No(50) \\
        AgNews & World(25), Sports(25), Business(25), Sci/Tech(25)\\
        COPA & Cause(50), Effect(50)\\
        Com2sense & Yes(50), No(50) \\
        TriviaQA & - \\
        Mathdataset & - \\
        \hline
    \end{tabular}
    \caption{Datasets statistics: labels and total number of samples per label in brackets.}
    \label{tab:data_statistics}
\end{table*}
\endgroup
\begin{tcolorbox}[colback=white, colframe=black!20!white, title=\textbf{Medical Question Pair (Classification task)}, rounded corners, coltitle=black]
\textcolor{cyan}{In this task you are given a medical question pair. Your task is to classify this question pair into two categories 1) 'Similar' if the given two questions have the same connotation or meaning  2) 'Dissimilar' if the given two questions have a different connotation or meaning.} \\\\
\textcolor{gray}{Question1: After how many hour from drinking an antibiotic can I drink alcohol? Question2: I have a party tonight and I took my last dose of Azithromycin this morning. Can I have a few drinks?} \textcolor{orange}{Are these two questions similar or dissimilar?} \textcolor{green}{Similar}. \\\\
\textcolor{gray}{Question1: After how many hour from drinking an antibiotic can I drink alcohol? Question2: I vomited this morning and I am not sure if it is the side effect of my antibiotic or the alcohol I took last night...} \textcolor{orange}{Are these two questions similar or dissimilar?} \textcolor{green}{Dissimilar}. \\\\
\textcolor{gray}{Question1: Can coarctation of the aorta cause poor growth in height? Question2: I am 4' 8". My mom said that I have a birth defect (coarctation of aorta). Are the two related?} \textcolor{orange}{Are these two questions similar or dissimilar?} \textcolor{green}{Similar}. \\\\
\textcolor{gray}{Question1: Aspirin allergy - is it worth getting a bracelet? Question2: How much Aspirin can I take for my headache without causing any side effects?} \textcolor{orange}{Are these two questions similar or dissimilar?} \textcolor{green}{Dissimilar}.\\\\
\textcolor{darkgray}{[Test instance.]} \textcolor{orange}{Are these two questions similar or dissimilar?}

\end{tcolorbox}

\begin{tcolorbox}[colback=white, colframe=black!20!white, title=\textbf{Twitter Emotion Classification (Classification task)}, rounded corners, coltitle=black]
 \textcolor{cyan}{In this task, you are given a tweet. The task is to classify this tweet based on its emotion. The answer should be one of these emotions 'Sadness', 'Joy', 'Love', 'Anger', 'Fear', or 'Surprise'.} \\\\
\textcolor{gray}{Im feeling quite sad and sorry for myself but ill snap out of it soon.} \textcolor{orange}{Which emotion is expressed in this tweet?} \textcolor{green}{Sadness}. \\\\
\textcolor{gray}{I am just feeling cranky and blue.} \textcolor{orange}{Which emotion is expressed in this tweet?} \textcolor{green}{Anger}.\\\\
\textcolor{gray}{I can have for a treat or if i am feeling festive.} \textcolor{orange}{Which emotion is expressed in this tweet?} \textcolor{green}{Joy}.\\\\
\textcolor{gray}{I feel like im caring about my body not in just an attempt to be the right size but to feel good and have a full life.}\textcolor{orange}{Which emotion is expressed in this tweet?} \textcolor{green}{Love}.\\\\
\textcolor{darkgray}{[Test instance.]} \textcolor{orange}{Which emotion is expressed in this tweet?}

\end{tcolorbox}

\begin{tcolorbox}[colback=white, colframe=black!20!white, title=\textbf{CoLA (Classification task)}, rounded corners, coltitle=black]
 \textcolor{cyan}{You will be given a sentence. If the sentence is grammatically correct and meaningful, then answer with 'Yes', otherwise 'No'.} \\\\
\textcolor{gray}{Our friends won't buy this analysis, let alone the next one we propose.} \textcolor{orange}{Is this sentence meaningful and grammatically correct?} \textcolor{green}{Yes}. \\\\
\textcolor{gray}{One more pseudo generalization and I'm giving up.} \textcolor{orange}{Is this sentence meaningful and grammatically correct?} \textcolor{green}{Yes}. \\\\
\textcolor{gray}{They drank the pub.} \textcolor{orange}{Is this sentence meaningful and grammatically correct?} \textcolor{green}{No}. \\\\
\textcolor{gray}{Day by day the facts are getting murkier.} \textcolor{orange}{Is this sentence meaningful and grammatically correct?} \textcolor{green}{Yes}.\\\\
\textcolor{darkgray}{[Test instance.]} \textcolor{orange}{Is this sentence meaningful and grammatically correct?}

\end{tcolorbox}

\begin{tcolorbox}[colback=white, colframe=black!20!white, title=\textbf{Com2sense (Classification task)}, rounded corners, coltitle=black]
 
\textcolor{cyan}{You will be given a piece of text either about an everyday event, or a general statement. If the event seems a plausible event, or the general statement makes sense to you then answer the question as 'Yes', otherwise 'No'.} \\\\
\textcolor{gray}{The glass fell of a three-story building, so it broke into pieces.} \textcolor{orange}{Does this statement make sense to you?} \textcolor{green}{Yes}. \\ \\
\textcolor{gray}{Marry was going out to work, so she asked her sixteen-year-old daughter to take care of her five-year-old son.} \textcolor{orange}{Does this statement make sense to you?} \textcolor{green}{Yes}. \\\\
\textcolor{gray}{Johnathan didn't have a hammer, so he used a cotton pad to drive the nail into the wood. }\textcolor{orange}{Does this statement make sense to you?}  \textcolor{green}{No}. \\\\
\textcolor{gray}{Suraya's best friend is getting married soon, so she will likely choose to go on a trip instead of helping her friend organize the ceremony.} \textcolor{orange}{Does this statement make sense to you?} \textcolor{green}{No}. \\\\
\textcolor{darkgray}{[Test instance.]} \textcolor{orange}{Does this statement make sense to you?}
\end{tcolorbox}

\begin{tcolorbox}[colback=white, colframe=black!20!white, title=\textbf{RTE (Classification Task)}, rounded corners, coltitle=black, fontupper=\small]
 
\textcolor{cyan}{In this task, you are given two sentences. Indicate if the first sentence clearly entails the second sentence (i.e., one can conclude the 2nd sentence by reading the 1st one). Indicate your answer with 'Yes' if the first sentence entails the second sentence, otherwise answer with 'No'.} \\\\
\textcolor{gray}{Sentence 1: No Weapons of Mass Destruction Found in Iraq Yet. Sentence 2:Weapons of Mass Destruction Found in Iraq.} \textcolor{orange}{Does Sentence 1 entail Sentence 2?} \textcolor{green}{No}. \\\\
\textcolor{gray}{Sentence 1: A place of sorrow, after Pope John Paul II died, became a place of celebration, as Roman Catholic faithful gathered in downtown Chicago to mark the installation of new Pope Benedict XVI. Sentence 2: Pope Benedict XVI is the new leader of the Roman Catholic Church.} \textcolor{orange}{Does Sentence 1 entail Sentence 2?} \textcolor{green}{Yes}. \\\\
\textcolor{gray}{Sentence 1: Herceptin was already approved to treat the sickest breast cancer patients, and the company said, Monday, it will discuss with federal regulators the possibility of prescribing the drug for more breast cancer patients. Sentence 2: Herceptin can be used to treat breast cancer.} \textcolor{orange}{Does Sentence 1 entail Sentence 2?} \textcolor{green}{Yes}. \\\\
\textcolor{gray}{Sentence 1: Nearly 4 million children who have at least one parent who entered the U.S. illegally were born in the United States and are U.S. citizens as a result, according to the study conducted by the Pew Hispanic Center. That's about three quarters of the estimated 5.5 million children of illegal immigrants inside the United States, according to the study. About 1.8 million children of undocumented immigrants live in poverty, the study found. Sentence 2: Three quarters of U.S. illegal immigrants have children.} \textcolor{orange}{Does Sentence 1 entail Sentence 2?} \textcolor{green}{No}. \\\\ 
\textcolor{darkgray}{[Test instance.]} \textcolor{orange}{Does Sentence 1 entail Sentence 2?}
  
\end{tcolorbox}

\begin{tcolorbox}[colback=white, colframe=black!20!white, title=\textbf{Financial Phrasebank (Classification task)}, rounded corners, coltitle=black]
 \textcolor{cyan}{Classify the given a piece of financial news into three classes: positive, negative, and neutral. Output must be 'Positive', 'Negative', or 'Neutral'.} \\\\
\textcolor{gray}{According to Gran , the company has no plans to move all production to Russia , although that is where the company is growing.} \textcolor{orange}{Is the sentiment of the sentence 'Negative', 'Neutral', or 'Positive'?} \textcolor{green}{Neutral}.\\\\
\textcolor{gray}{Technopolis plans to develop in stages an area of no less than 100,000 square meters in order to host companies working in computer technologies and telecommunications , the statement said.} \textcolor{orange}{Is the sentiment of the sentence 'Negative', 'Neutral', or 'Positive'?} \textcolor{green}{Neutral}.\\\\
\textcolor{gray}{The international electronic industry company Elcoteq has laid off tens of employees from its Tallinn facility ; contrary to earlier layoffs the company contracted the ranks of its office workers , the daily Postimees reported.} \textcolor{orange}{Is the sentiment of the sentence 'Negative', 'Neutral', or 'Positive'?} \textcolor{green}{Negative}.\\\\
\textcolor{gray}{With the new production plant the company would increase its capacity to meet the expected increase in demand and would improve the use of raw materials and therefore increase the production profitability.} \textcolor{orange}{Is the sentiment of the sentence 'Negative', 'Neutral', or 'Positive'?} \textcolor{green}{Positive}.\\\\
\textcolor{darkgray}{[Test instance.]}  \textcolor{orange}{Is the sentiment of the sentence 'Negative', 'Neutral', or 'Positive'?}
\end{tcolorbox}

\begin{tcolorbox}[colback=white, colframe=black!20!white, title=\textbf{Mathdataset Answer Generation (Generation task)}, rounded corners, coltitle=black]
 \textcolor{cyan}{Given a simple high-school level math question, you are required to solve it and provide the final answer. The final answer is always a single number. These questions can range from a variety of topics like simple arithmetic, solving equations, converting a quantity from one unit to another, finding remainders/GCD/LCM, finding probabilities etc. Each question has only one correct answer. This answer can be a positive or negative integer, a fraction or a decimal number. If the answer is a negative number use the hyphen (e.g. -42) symbol for the minus sign. For decimal numbers, do not add extra zeros after the decimal point. For fractional numbers, separate the numerator and denominator using a forward slash (e.g. 3/25).} \\\\
\textcolor{gray}{Let y = -74 - -79. Solve 0 = -y*q - 13 + 3 for q.} \textcolor{orange}{The answer to this math problem is} \textcolor{green}{-2}. \\\\
\textcolor{gray}{Work out 29.8 + -0.18.} \textcolor{orange}{The answer to this math problem is} \textcolor{green}{29.62}.\\\\
\textcolor{gray}{How many nanometers are there in 610.1077 millimeters} \textcolor{orange}{The answer to this math problem is} \textcolor{green}{610107700}.\\\\
\textcolor{gray}{Four letters picked without replacement from bboobleoeewobw. What is prob of picking 3 o and 1 e?} \textcolor{orange}{The answer to this math problem is} \textcolor{green}{12/1001}.\\\\
\textcolor{darkgray}{[Test instance.]} \textcolor{orange}{The answer to this math problem is}
\end{tcolorbox}

\begin{tcolorbox}[colback=white, colframe=black!20!white, title=\textbf{AGNews (Classification task)}, rounded corners, coltitle=black]
 \textcolor{cyan}{In this task, you are given a news article. Your task is to classify the article to one out of the four topics 'World', 'Sports', 'Business', 'Sci/Tech'. If you are not sure about the topic, choose the closest option. Note that URLs in the text have been replaced with [Link].} \\\\
Comets, Asteroids and Planets around a Nearby Star (SPACE.com) SPACE.com - A nearby star thought to harbor comets and asteroids now appears to be home to planets, too. The presumed worlds are smaller than Jupiter and could be as tiny as Pluto, new observations suggest. \textcolor{orange}{What label best describes this news article?} Sci/Tech. \\\\
\textcolor{gray}{Oil and Economy Cloud Stocks' Outlook  NEW YORK (Reuters) - Soaring crude prices plus worries  about the economy and the outlook for earnings are expected to  hang over the stock market next week during the depth of the  summer doldrums.} \textcolor{orange}{What label best describes this news article?} \textcolor{green}{Business}.\\\\
\textcolor{gray}{Russian FM meets with Katsav Russian Foreign Minister Sergey Lavrov met Monday with Israeli 39;s President Moshe Katsav as part of his first tour of the region to discuss, among other topics, a collaboration between the two countries in combating terrorism.} \textcolor{orange}{What label best describes this news article?} \textcolor{green}{World}. \\\\
\textcolor{gray}{Murtagh a stickler for success Northeastern field hockey coach Cheryl Murtagh doesn't want the glare of the spotlight that shines on her to detract from a team that has been the America East champion for the past three years and has been to the NCAA tournament 13 times.} \textcolor{orange}{What label best describes this news article?} \textcolor{green}{Sports}. \\\\
\textcolor{darkgray}{[Test instance.]} \textcolor{orange}{What label best describes this news article?}
\end{tcolorbox}

\begin{tcolorbox}[colback=white, colframe=black!20!white, title=\textbf{COPA (Classification task)}, rounded corners, coltitle=black]
 \textcolor{cyan}{In this task your given two statements. You must judge whether the second sentence is the cause or effect of the first sentence.  The two sentences are separated by a newline character and the answer can be 'Cause' or 'Effect'.} \\\\
\textcolor{gray}{The women met for coffee.
They wanted to catch up with each other.} \textcolor{orange}{Is the second sentence cause or effect of the first sentence?} \textcolor{green}{Cause}. \\\\
\textcolor{gray}{The physician misdiagnosed the patient.
The patient filed a malpractice lawsuit against the physician.} \textcolor{orange}{Is the second sentence cause or effect of the first sentence?} \textcolor{green}{Effect}.\\\\
\textcolor{gray}{The guests of the party hid behind the couch.
It was a surprise party.} \textcolor{orange}{Is the second sentence cause or effect of the first sentence?} \textcolor{green}{Cause}.\\\\
\textcolor{gray}{My friend was recovering from surgery.
I brought her a card and flowers.} \textcolor{orange}{Is the second sentence cause or effect of the first sentence?} \textcolor{green}{Effect}. \\\\
\textcolor{darkgray}{[Test instance.]} \textcolor{orange}{Is the second sentence cause or effect of the first sentence?}
\end{tcolorbox}

\begin{tcolorbox}[colback=white, colframe=black!20!white, title=\textbf{TriviaQA (Generation task)}, rounded corners, coltitle=black]
 \textcolor{cyan}{You are given a general knowledge question based on Wikipedia and Web content. Write an answer to this question.} \\\\
\textcolor{gray}{Who was the man behind The Chipmunks?} \textcolor{orange}{The answer to this question is} \textcolor{green}{David Seville}.\\\\
\textcolor{gray}{What star sign is Jamie Lee Curtis?} \textcolor{orange}{The answer to this question is} \textcolor{green}{Scorpio.} \\\\
\textcolor{gray}{Which Lloyd Webber musical premiered in the US on 10th December 1993?} \textcolor{orange}{The answer to this question is} \textcolor{green}{Sunset Boulevard}.\\\\
\textcolor{gray}{The Euro is divided into how many cents?} \textcolor{orange}{The answer to this question is} \textcolor{green}{100}.\\\\
\textcolor{darkgray}{[Test instance.]} \textcolor{orange}{The answer to this question is}
\end{tcolorbox}



\section{More results}

Table \ref{tab:avg_result_across_class_n_gen_tasks} shows results for classification and generation tasks whereas Tables \ref{tab:avg_result_across_dataset_tasks_gpt2-xl} and \ref{tab:avg_result_across_dataset_tasks_llama-70b} show results for individual tasks for GPT-2-xl (smaller model) and Llama-2-70B (bigger model) respectively.

\begingroup
\begin{table*}
    \centering
    \begin{tabular}{ p{0.6\columnwidth} | r r r r}
    \toprule
       Corruptions & GPT2-xl & GPT2-xl &  LLama-2-70B & LLama-2-70B \\
        & Classification  & Generation  &  Classification  & Generation  \\
        \midrule   
        Semantic Corruptions & & & & \\
        + demos. & 28.1 & 9.8 & 53.8 & 30.7 \\
        + task instr. + demos. & 30.4 & 11.2 & 61.6 & 33.0 \\
        + inline instr. + demos. & 38.3 & 9.3 & 69.0 & 32.5 \\
        \midrule
        Baseline & \underline{40.7} & \underline{\textbf{15.5}} & \underline{72.3} & \underline{33.4} \\
        \midrule
        Semantic Corruption & & & & \\
        Rw both instr. & 40.4 & 11.6 & 52.1 & 30.8 \\
        Rw labels & 3.7 & 6.3 & 1.1 & 1.7 \\
        OOD inputs & 41.7 & 13.3 & 57.6 & 22.1 \\
        \midrule
        Repeated text & & & & \\
        Inline instr. in 3 demos & \textbf{45.3} & 15.1 & \textbf{72.4} & \textbf{33.5} \\
        Inline instr. in 2 demos & 41.4 & 14.7 & 70.4 & 31.9 \\
        Inline instr. in 1 demos & 41.4 & 15.3 & 69.6 & 31.7 \\
        Inline instr. in 0 demos & 17.6 & 9.1 & 38.1 & 22.9 \\
        Rw Inline instr. in 3 demos & 41.1 & 10.2 & 53.5 & 27.0 \\
        Rw Inline instr. in 2 demos & 41.6 & 9.1 & 54.0 & 29.8 \\
        Rw Inline instr. in 1 demos & 40.2 & 3.0 & 41.5 & 22.5 \\
        Rw Inline instr. in 0 demos & 0.9 & 0.0 & 0.9 & 0.2 \\        
        \bottomrule
    \end{tabular}
\caption{\footnotesize Model performance for classification and generation tasks. The highest performance is in bold, baseline prompt performance is underlined.}
\label{tab:avg_result_across_class_n_gen_tasks}
\end{table*}
\endgroup

\begingroup
\begin{table*}
\begin{adjustbox}{width=\textwidth}
    \centering
    \begin{tabular}{ p{0.55\columnwidth} | r r r r r r r r r r}
    \toprule
     \textbf{Corruption} & \textbf{RTE} & \textbf{MQP} & \textbf{FPH} & \textbf{TE} & \textbf{CoLA} & \textbf{AGN} & \textbf{COPA} & \textbf{C2S} & \textbf{TQ} & \textbf{MATH} \\ 
        \midrule
        Structural & & & & & & & & & & \\
        + demos. & 46.0 & 46.0 & 19.0 & 6.0 & 49.0 & 33.0 & 20.0 & 42.0 & 18.6 & 0.9 \\
        + task instr. & 45.0 & 47.0 & 25.0 & 19.0 & 48.0 & 39.0 & 22.0 & 37.0 & 20.2 & 2.1 \\
        + inline instr. & 50.0 & 50.0 & 35.0 & 26.0 & 50.0 & 61.0 & 54.0 & 50.0 & 15.1 & 3.6 \\
        \midrule
        Baseline & 53.0 & 60.0 & 34.0 & 23.0 & 51.0 & 58.0 & 50.0 & 47.0 & 17.0 & 14.0 \\
        \midrule
        Semantics & & & & & & & & & & \\
        Rw both instr. & 53.0 & 50.0 & 58.0 & 25.0 & 44.0 & 54.0 & 50.0 & 47.0 & 15.6 & 7.7 \\
        Rw labels & 0.0 & 0.0 & 0.0 & 23.0 & 0.0 & 1.0 & 0.0 & 0.0 & 11.9 & 0.7 \\
        OOD inputs & 51.0 & 52.0 & 46.0 & 28.0 & 50.0 & 63.0 & 52.0 & 48.0 & 12.7 & 13.9 \\
        \midrule
        Repeated text & & & & & & & & & & \\
        Inline instr. in 3 demos & 52.0 & 63.0 & 50.0 & 22.0 & 50.0 & 41.0 & 50.0 & 51.0 & 18.0 & 12.2 \\
        Inline instr. in 2 demos & 54.0 & 70.0 & 35.0 & 21.0 & 50.0 & 32.0 & 50.0 & 50.0 & 14.9 & 14.6 \\
        Inline instr. in 1 demos & 50.0 & 50.0 & 33.0 & 31.0 & 50.0 & 38.0 & 50.0 & 50.0 & 15.1 & 15.5 \\
        Inline instr. in 0 demos & 1.0 & 43.0 & 0.0 & 1.0 & 47.0 & 16.0 & 0.0 & 0.0 & 9.1 & 9.3 \\
        Rw Inline instr. in 3 demos & 50.0 & 50.0 & 40.0 & 20.0 & 42.0 & 37.0 & 50.0 & 49.0 & 15.4 & 5.0 \\
        Rw Inline instr. in 2 demos & 50.0 & 50.0 & 33.0 & 20.0 & 50.0 & 35.0 & 53.0 & 50.0 & 11.5 & 6.8 \\
        Rw Inline instr. in 1 demos & 50.0 & 50.0 & 33.0 & 16.0 & 50.0 & 25.0 & 50.0 & 50.0 & 0.0 & 6.0 \\
        Rw Inline instr. in 0 demos & 0.0 & 0.0 & 0.0 & 0.0 & 4.0 & 0.0 & 0.0 & 2.0 & 0.0 & 0.0 \\
        \bottomrule
    \end{tabular}
\end{adjustbox}
\caption{\footnotesize \textbf{Model performance for each dataset for GPT2-xl}. 
Datasets are RTE, Medical Question Pair (MQP), Financial Phrasebank (FPH), Twitter Emotion classification(TE), CoLA, AgNews (AGN), COPA, Com2sense (C2S), and two generation tasks: TriviaQA (TQ) and Mathdataset answer generation(MATH)
}
\label{tab:avg_result_across_dataset_tasks_gpt2-xl}
\end{table*}
\endgroup

\begingroup
\begin{table*}
\begin{adjustbox}{width=\textwidth}
    \centering
    \begin{tabular}{ p{0.55\columnwidth} | r r r r r r r r r r}
    \toprule
     \textbf{Corruption} & \textbf{RTE} & \textbf{MQP} & \textbf{FPH} & \textbf{TE} & \textbf{CoLA} & \textbf{AGN} & \textbf{COPA} & \textbf{C2S} & \textbf{TQ} & \textbf{MATH} \\ 
        \midrule
        Structural & & & & & & & & & & \\
        + demos. & 61.0 & 60.0 & 60.0 & 22.0 & 49.0 & 58.0 & 67.0 & 53.0 & 41.6 & 19.8 \\
        + task instr. & 67.0 & 66.0 & 69.0 & 22.0 & 72.0 & 61.0 & 64.0 & 72.0 & 41.8 & 24.2 \\
        + inline instr. & 70.0 & 67.0 & 78.0 & 42.0 & 72.0 & 83.0 & 75.0 & 65.0 & 42.4 & 22.2 \\
        \midrule
        Baseline & 84.0 & 77.0 & 80.0 & 34.0 & 81.0 & 86.0 & 64.0 & 72.0 & 42.5 & 24.3 \\
        \midrule
        Semantics & & & & & & & & & & \\
        Rw both instr. & 66.0 & 53.0 & 34.0 & 27.0 & 54.0 & 74.0 & 56.0 & 56.0 & 42.9 & 18.7 \\
        Rw labels & 3.0 & 0.0 & 0.0 & 1.0 & 0.0 & 0.0 & 0.0 & 5.0 & 1.7 & 1.7 \\
        OOD inputs & 70.0 & 61.0 & 71.0 & 20.0 & 77.0 & 35.0 & 54.0 & 73.0 & 35.0 & 9.2 \\
        \midrule
        Repeated text & & & & & & & & & & \\
        Inline instr. in 3 demos & 76.0 & 82.0 & 84.0 & 44.0 & 74.0 & 85.0 & 63.0 & 71.0 & 45.8 & 21.2 \\
        Inline instr. in 2 demos & 72.0 & 80.0 & 78.0 & 41.0 & 72.0 & 86.0 & 65.0 & 69.0 & 40.4 & 23.3 \\
        Inline instr. in 1 demos & 70.0 & 81.0 & 82.0 & 37.0 & 76.0 & 84.0 & 60.0 & 67.0 & 43.2 & 20.3 \\
        Inline instr. in 0 demos & 43.0 & 52.0 & 34.0 & 24.0 & 71.0 & 1.0 & 25.0 & 55.0 & 25.3 & 20.5 \\
        Rw Inline instr. in 3 demos & 63.0 & 55.0 & 40.0 & 35.0 & 49.0 & 79.0 & 58.0 & 49.0 & 39.5 & 14.5 \\
        Rw Inline instr. in 2 demos & 58.0 & 70.0 & 33.0 & 34.0 & 50.0 & 77.0 & 59.0 & 51.0 & 39.9 & 19.7 \\
        Rw Inline instr. in 1 demos & 50.0 & 58.0 & 34.0 & 18.0 & 48.0 & 25.0 & 50.0 & 49.0 & 39.1 & 6.0 \\
        Rw Inline instr. in 0 demos & 5.0 & 0.0 & 0.0 & 0.0 & 1.0 & 1.0 & 0.0 & 0.0 & 0.4 & 0.0 \\
        \bottomrule
    \end{tabular}
\end{adjustbox}
\caption{\footnotesize \textbf{Model performance for each dataset for LLama-2-70B}. 
Datasets are RTE, Medical Question Pair (MQP), Financial Phrasebank (FPH), Twitter Emotion classification(TE), CoLA, AgNews (AGN), COPA, Com2sense (C2S), and two generation tasks: TriviaQA (TQ) and Mathdataset answer generation(MATH)
}
\label{tab:avg_result_across_dataset_tasks_llama-70b}
\end{table*}
\endgroup

\section{Attention plots}
\label{appendix_attention_plots}

\begin{figure*}[h!]
    \centering
    \includegraphics[scale=0.5]{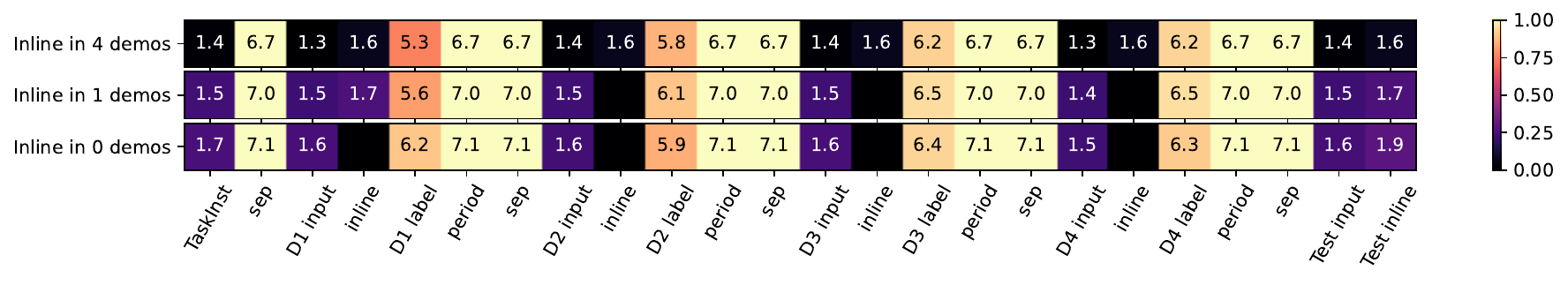}
    \caption{Average GPT-J-6B attention per component for repeated text corruptions. ``Inline'' refers to the presence of the number of inline instructions in the baseline prompt. Fully black box represents missing components. }
    \label{fig:gptj_repeatedtextcorrup_attention}
\end{figure*}

\begin{figure*}
    \centering
    \includegraphics[scale=0.5]{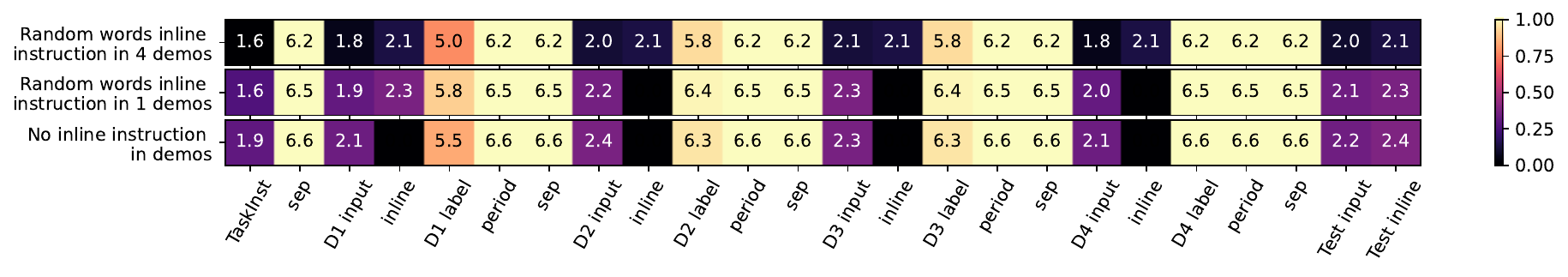}
    \caption{ Random words instructions: Average OPT-30B attention per component for repeated text corruptions. ``Inline'' refers to the presence of the number of inline instructions in the baseline prompt. A solid black box represents omitted components. }
    \label{fig:opt-30b_repeatedtextcorrup_attention_rw}
\end{figure*}

Figure \ref{fig:gptj_repeatedtextcorrup_attention} shows repeated text corruptions for GPT-J-6B. Figure \ref{fig:opt-30b_repeatedtextcorrup_attention_rw} shows repeated text corruptions for OPT-30B with random word instructions.


\end{document}